\documentclass[10pt]{article}
\usepackage[utf8]{inputenc}
\usepackage[T1]{fontenc}
\usepackage{absspconf,amsmath}
\usepackage{tikz}
\usepackage{amssymb}
\usepackage{tabularx}
\usepackage{multirow}
\usepackage{siunitx}
\usepackage{pifont}
\usepackage{graphicx}
\usepackage{lipsum}
\usepackage{amsfonts}
\usepackage{booktabs}
\usepackage{blindtext}
\usepackage{csquotes}
\usepackage{makecell}
\usepackage{tikz}
\usepackage{mathtools}
\usepackage{overpic}
\usepackage{tikzscale}
\usepackage{gnuplot-lua-tikz}
\usepackage{subcaption}
\usepackage{microtype}
\usepackage{float}
\usepackage[above]{placeins}

\usepackage{array}
\usepackage[style=ieee, doi=false, url=false, isbn=false, minnames=1, maxnames=8, date=year, hyperref=true]{biblatex}
\defbibheading{secbib}[\bibname]{%
  \section{#1}%
  \markboth{#1}{#1}
}
\renewbibmacro{in:}{}
\DeclareBibliographyDriver{unpublished}{%
    \usebibmacro{bibindex}%
    \usebibmacro{begentry}%
    \usebibmacro{author/translator+others}%
    \newunit
    \usebibmacro{title}%
    \newunit\newblock
    \usebibmacro{eprint}%
    \newunit\newblock
    \usebibmacro{date}
    \usebibmacro{finentry}%
}
\DeclareFieldFormat{eprint:arxiv}{%
  arXiv\addcolon\space
  \ifhyperref
    {\href{https://arxiv.org/abs/#1}{
       \nolinkurl{#1}%
       \iffieldundef{eprintclass}
         {}
         }}
    {
    #1
     \iffieldundef{eprintclass}
       {}
       }
}
\usepackage{caption, setspace}
\usepackage{soul}
\usepackage[acronym]{glossaries}
\usepackage{etoolbox}

\usepackage{comment}
\usepackage{todonotes}
\usepackage{url}
\usepackage[hidelinks]{hyperref}
\usepackage[capitalise]{cleveref} 

\makeatletter
\let\oldmakefirstuc\makefirstuc
\renewcommand*{\makefirstuc}[1]{%
  \def\gls@add@space{}%
  \mfu@capitalisewords#1 \@nil\mfu@endcap
}
\def\mfu@capitalisewords#1 #2\mfu@endcap{%
  \def\mfu@cap@first{#1}%
  \def\mfu@cap@second{#2}%
  \gls@add@space
  \oldmakefirstuc{#1}%
  \def\gls@add@space{ }%
  \ifx\mfu@cap@second\@nnil
    \let\next@mfu@cap\mfu@noop
  \else
    \let\next@mfu@cap\mfu@capitalisewords
  \fi
  \next@mfu@cap#2\mfu@endcap
}
\makeatother

\title{Estimating Physical Information Consistency of Channel Data Augmentation for Remote Sensing Images}
\name{Tom Burgert$^{1,2}$, Beg\"{u}m Demir$^{1,2}$
 }
\address{
    $^1$BIFOLD -- Berlin Institute for the Foundations of Learning and Data, Germany\\%
    $^2$Faculty of Electrical Engineering and Computer Science, Technische Universität Berlin, Germany
}

\begin{document}

\maketitle

\begin{abstract}
The application of data augmentation for deep learning (DL) methods plays an important role in achieving state-of-the-art results in supervised, semi-supervised, and self-supervised image classification. In particular, channel transformations (e.g., solarize, grayscale, brightness adjustments) are integrated into data augmentation pipelines for remote sensing (RS) image classification tasks. However, contradicting beliefs exist about their proper applications to RS images. A common point of critique is that the application of channel augmentation techniques may lead to physically inconsistent spectral data (i.e., pixel signatures). To shed light on the open debate, we propose an approach to estimate whether a channel augmentation technique affects the physical information of RS images. To this end, the proposed approach estimates a score that measures the alignment of a pixel signature within a time series that can be naturally subject to deviations caused by factors such as acquisition conditions or phenological states of vegetation. We compare the scores associated with original and augmented pixel signatures to evaluate the physical consistency. Experimental results on a multi-label image classification task show that channel augmentations yielding a score that exceeds the expected deviation of original pixel signatures can not improve the performance of a baseline model trained without augmentation.
\glsresetall
\end{abstract}

\begin{keywords}
Data augmentation, physically consistent data, multi-label image classification, deep learning, remote sensing.
\end{keywords}

\begin{figure*}
    \renewcommand{\arraystretch}{3}
    \centering
        \begin{overpic}[width=.95\linewidth]{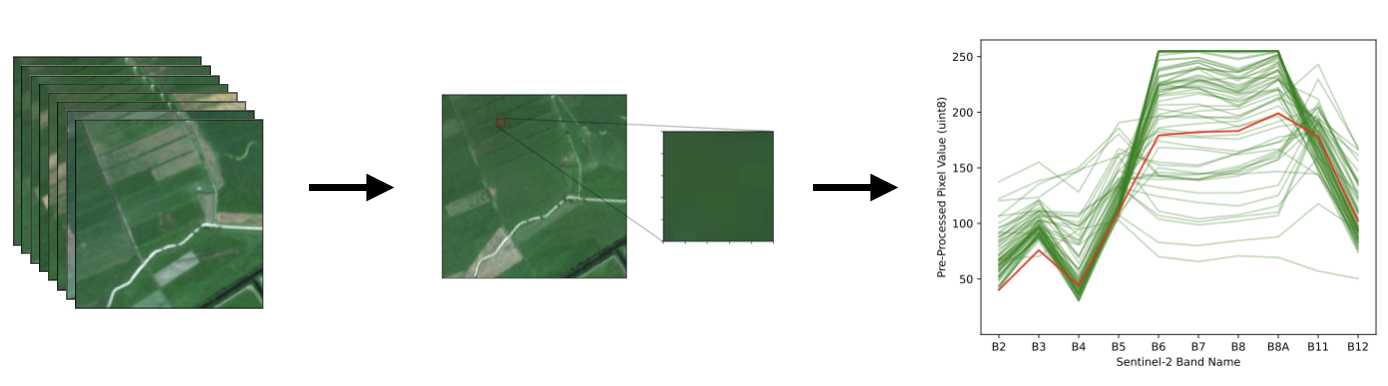}
            \small
            \put(8.8, -2.5){\textbf{Step 1}}
            \put(42.5, -2.5){\textbf{Step 2}}
            \put(82.4, -2.5){\textbf{Step 3}}
        \end{overpic}\\[1.5em] 
    \caption{Qualitative overview for calculating $S_{\text{aug}}$ (or $S_{\text{noaug}}$). Step 1: Construct a set $D$ of $N$ time series $\mathbf{t}_i$. In the case of optical data, filter out all cloudy images. Step 2: Per time series $\mathbf{t}_i$ select a homogeneous $k \times k$ sub-area that is not impacted by any land use land cover changes over time to generate mask $\mathbf{b}_i$. Step 3: For each image $\mathbf{t}_{i,\tau}$ in $\mathbf{t}_i$ compute the respective pixel signature by averaging over the selected sub-area per band. Then, for each signature (red) from an augmented image for $S_{\text{aug}}$ (or unaugmented image for $S_{\text{noaug}}$), compute the distance to the closest unaugmented signature (green) within the same time series.}
    \label{fig:method_overview}
\end{figure*}

\section{Introduction}

The development of deep learning (DL) based methods for remote sensing (RS) image classification has recently gained increasing attention in the context of single-label \cite{lu_rsi-mix_2022}, multi-label \cite{burgert_effects_2022}, and pixel-based classification \cite{pan_coinnet_2019}. Due to costly and time-consuming processes of collecting labeled training data, the development of data augmentation techniques is a growing research interest to improve the generalization capabilities of DL methods. Data augmentation refers to the process of creating slightly modified versions of existing training images by applying stochastic transformations to them while preserving their semantic characteristics \cite{burgert_label_2024}. Data augmentation techniques play a central role not only for state-of-the-art results in supervised learning \cite{liu_convnet_2022} but also for semi-supervised \cite{sohn_fixmatch_2020} and self-supervised learning \cite{he_momentum_2020}. However, the accurate usage of data augmentation techniques for image classification tasks in RS is still an open debate. While geometric image transformations such as cropping, rotation, translation, or flipping are commonly used in data augmentation pipelines \cite{lu_rsi-mix_2022}, \cite{zhang_new_2021}, \cite{stivaktakis_deep_2019}, there exist ambiguous beliefs about the application of channel transformations. Patnala et al. \cite{patnala_generating_2023} assume that the application of channel augmentation techniques that distort the pixel signatures (i.e., the spectral data) affects the underlying physical information, resulting in physically inconsistent augmented data when used in contrastive self-supervised learning (SSL) for multispectral images. Consequently, they propose to avoid any type of channel augmentation technique. Contradicting this hypothesis, there exist multiple works in RS that employ the standard set of contrastive SSL augmentation techniques from the computer vision literature when training an SSL algorithm on multispectral images \cite{wang_ssl4eo-s12_2022}, \cite{zhao_hyperspectral_2022}, \cite{wang_self-supervised_2022}. These techniques include channel augmentations like brightness and contrast adjustments, grayscale, solarize, or Gaussian blur. In RS, pixel signatures of a given land cover class within a small area are naturally subject to small deviations at different image acquisition timestamps due to: i) distinct acquisition conditions (i.e., illumination, atmosphere, viewing angles, sensor parameters) that affect the radiometry of a scene; and ii) the phenological state of vegetation or the differences in the soil moisture that can lead to crucial variations in the spectral response of the same land cover classes (e.g. bare soil, crops). To shed light on the open debate on whether channel augmentation techniques can be considered as physical information consistent with these natural deviations, we propose an approach to measure the expected deviation (score) of pixel signatures at different timestamps. To this end, the proposed approach compares the expected deviation of an original (i.e., unaugmented) pixel signature with an augmented pixel signature within a time series. Experimental results on a multi-label multispectral image classification task show that channel augmentations yielding a score that exceeds the expected deviation of original pixel signatures can not improve the performance of a baseline model trained without augmentation.

\section{Methodology}

Let $\mathcal{D} \coloneqq \{\mathbf{t}_1, \cdots, \mathbf{t}_N\}$ be a set of $N$ time series, where $\mathbf{t}_i$ is the $i$-th time series defined as $(\mathbf{t}_{i,1} \cdots \mathbf{t}_{i,T})$ of length $T$ and $\mathbf{t}_{i,\tau}$ is the $\tau$-th image in $\mathbf{t}_i$. We assume each image $\mathbf{t}_{i,\tau} \in \mathbb{R}^{C \times H \times W}$ with a number of $C$ image bands, a height $H$ and width $W$ to be pre-processed to \textit{uint8}-values. Each time series $\mathbf{t}_i$ is associated with a binary mask $\mathbf{b}_i = \{0, 1\}^{C \times H \times W}$. The mask $\mathbf{b}_i$ contains an area of $k \times k$ pixels with ones not impacted by land cover changes over time and the rest of the masks filled with zeros. We define a function to extract the averaged signature of the $k \times k$-pixel area by $\text{sig}(\mathbf{x}, \mathbf{b}) \coloneqq \frac{\mathbf{x} \cdot \mathbf{b}}{\|\mathbf{b}\|_1}$. Let $\text{aug}(\cdot)$ represent a stochastic channel transformation with a given magnitude. Then, $\Tilde{\mathbf{t}}_{i,\tau} = \text{aug}(\mathbf{t}_{i,\tau})$ is defined as an augmented image within time series $\mathbf{t}_i$. In this paper, we propose an approach to measure whether the application of a channel augmentation technique to an image affects the physical information of its pixel signatures. Pixel signatures can be naturally subject to small deviations within a time series due to different acquisition conditions or phenological states of vegetation. The aim of this paper is to analyze whether the impact of channel augmentation techniques applied to an image yields pixel signatures that stay within these expected natural deviations. Therefore, we define a measure for the deviation of a pixel signature within a time series by

\begin{align}
    d(\mathbf{t}_{i,\tau}, \mathbf{t}_i, \mathbf{b}_i) \coloneqq \frac{1}{C} \min_{\substack{1 \leq {\tau}' \leq T \\ \tau \neq {\tau}'}} \| \text{sig}(\mathbf{t}_{i,\tau}, \mathbf{b}_i) - \text{sig}(\mathbf{t}_{i,\tau'}, \mathbf{b}_i) \|_1.
\end{align}

Thus, the score for the expected deviation of an original (i.e., unaugmented) pixel signature within a time series is estimated as follows:

\begin{align}
    S_{\text{noaug}} = \frac{1}{NT} \sum_{i=1}^{N} \sum_{\tau=1}^{T}  d(\mathbf{t}_{i,\tau}, \mathbf{t}_i, \mathbf{b}_i).
\end{align}

The score for the expected deviation of an augmented pixel signature within a time series is estimated as follows:

\begin{align}
    S_{\text{aug}} = \frac{1}{NTM} \sum_{i=1}^{N} \sum_{\tau=1}^{T} \sum_{m=1}^{M} d(\Tilde{\mathbf{t}}_{i,\tau}, \mathbf{t}_i, \mathbf{b}_i),
\end{align}

\noindent where $M$ represents the number of repetitions to account for the stochastic properties of the channel augmentation technique $\text{aug}(\cdot)$. The term augmented pixel signature refers to the signature extracted from an augmented version of the original image.

A qualitative overview of the calculation of the scores can be found in \cref{fig:method_overview}. First, $N$ time series $\mathbf{t}_i$ are constructed. In the case optical data is used, images with clouds are filtered out (Step 1). Even though standard registration methods can provide high overall registration accuracy, they may cause significant residual misalignment locally (e.g., spatially corresponding pixels belonging to different objects at two timestamps). To reduce the effect of local residual misalignment on the time series of an individual pixel signature, for each time series $\mathbf{t}_i$ a homogeneous $k \times k$-pixel sub-area that is not impacted by any land use land cover changes over time is manually selected to generate mask $\mathbf{b}_i$ (Step 2). For each image $\mathbf{t}_{i,\tau}$ in $\mathbf{t}_i$, the respective pixel signature is computed by averaging over the selected sub-area extracted by $\mathbf{b}_i$. Then, for each signature (red) from an unaugmented image (for $S_{\text{noaug}}$) or from an augmented image (for $S_{\text{aug}}$), the distance to the closest unaugmented signature (green) within the same time series is computed (Step 3). The final score is the expectation over $N$ time series.

\section{Experimental Results}

The experiments were conducted on the BigEarthNet-S2 dataset that includes Sentinel-2 multispectral images \cite{sumbul_bigearthnet_2019}. In detail, we have used the training set of the Lithuania split, while validation and test sets are defined using the Ireland split of this dataset. The training set contains $28226$ images, while each of the validation and test sets is composed of $12013$ images. Each training image is annotated with multi-labels. To compute the scores $S_{\text{noaug}}$ and $S_{\text{aug}}$, we have randomly selected $70$ images from the training set and then downloaded the respective time series between the years 2017 and 2020 to create $D$. We only consider images with one present class to ensure that any selected pixel in the image belongs to the same class. The classes represented in $D$ include arable land, broad-leaved forest, coniferous forest, inland waters, marine waters, pastures, urban fabric, while each filtered time series consists of $70$ to $100$ cloud-free images. We set the size of the sub-area $k \times k$ to $5 \times 5$ to, on the one hand, maximize the intersection over union of the spatial area among all images of the time series while, on the other hand, guaranteeing homogeneous pixel values. We calculate the score for the expected deviation of an augmented pixel signature within a time series $S_{\text{aug}}$ for the channel augmentation techniques of brightness, contrast, Gaussian blur, Gaussian noise, grayscale, posterize, sharpness and solarize. For a detailed description of the techniques, we refer the reader to Table 6 given in the appendix of \cite{cubuk_autoaugment_2019}. Note that some channel augmentation techniques require \textit{uint8}-values. Therefore, we follow a common image pre-processing strategy for Sentinel-2 images and divide the original \textit{uint16}-values by the 99th percentile per channel, followed by a 0-1-clipping and a multiplication by 255. We apply the pseudo-inverse of this strategy to the final scores to interpret them in the original \textit{uint16}-space. Except for the channel augmentation technique grayscale, we define a range of a maximum magnitude and subdivide the scale into $20$ bins for each technique (a maximum magnitude $\alpha_{\text{max}}$ of $20$ is associated with a strong transformation). Analogously to the application of augmentation techniques during training, we set the application probability of the augmentation techniques to $50$\% and sample a magnitude from the interval of a minimum magnitude of either 0 or $-\alpha_{\text{max}}$ and the maximum magnitude of $\alpha_{\text{max}}$ when calculating $S_{\text{aug}}$. Per augmentation technique and maximum magnitude, we set $M$ repetitions to $100$ to account for the stochastic properties of the channel augmentation technique. 

\begin{figure*}[t]
\centering
    \captionsetup[subfigure]{margin=-0.3cm}
    \def\subfigfrac{.24}
    \begin{subfigure}{\subfigfrac\linewidth}
        \includegraphics[width=\linewidth]{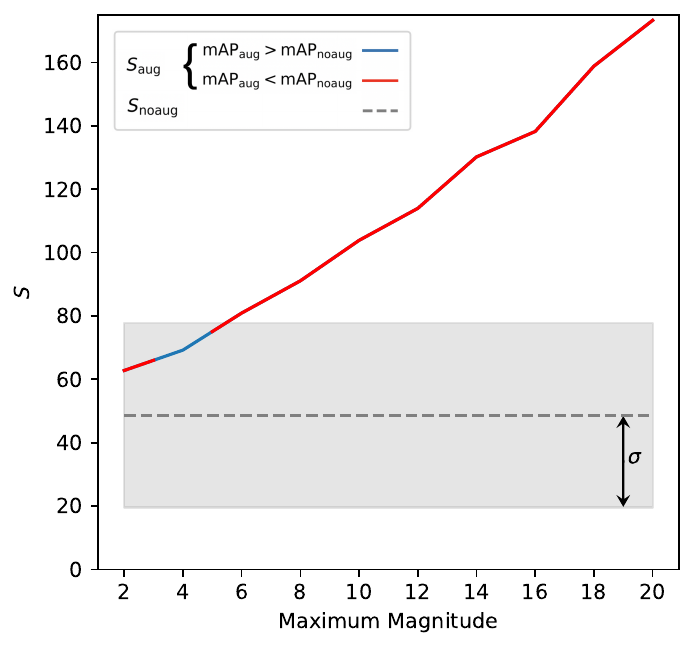}\caption{}\label{subfigure:seg_noise_original}
    \end{subfigure}
    \begin{subfigure}{\subfigfrac\linewidth}
        {\includegraphics[width=\linewidth]{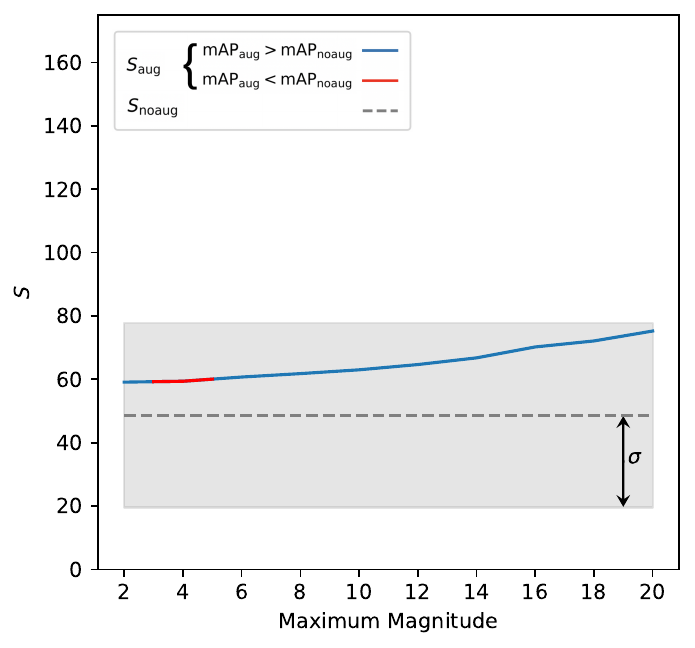}}\caption{}\label{subfigure:seg_noise_original_seg}
    \end{subfigure}
    \begin{subfigure}{\subfigfrac\linewidth}
        {\includegraphics[width=\linewidth]{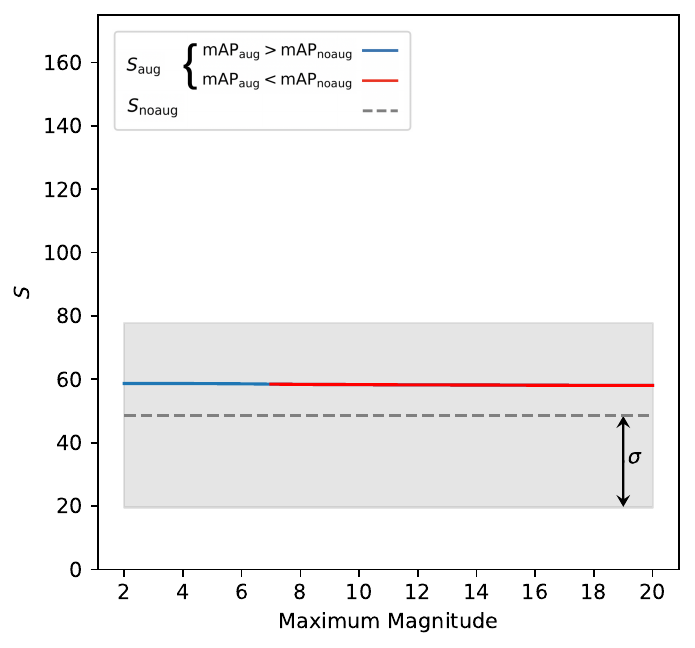}}\caption{}\label{subfigure:seg_noise_shift}
    \end{subfigure}
    \begin{subfigure}{\subfigfrac\linewidth}
        {\includegraphics[width=\linewidth]{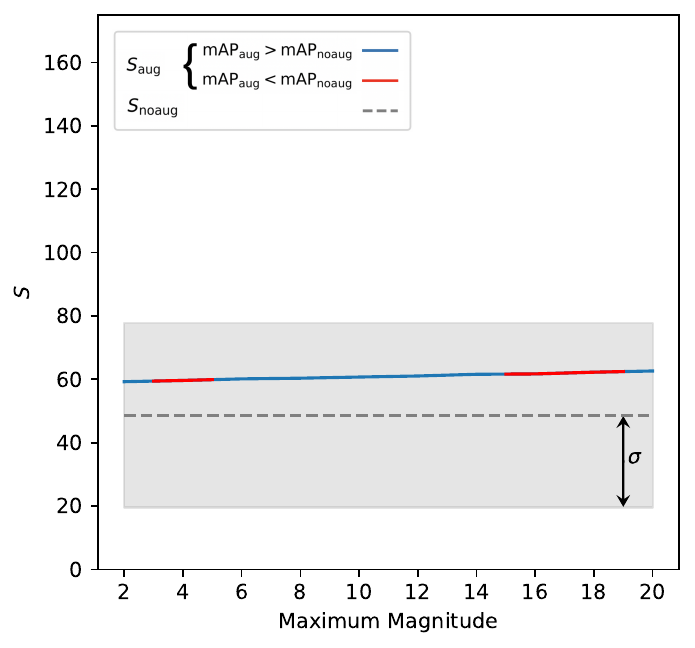}}\caption{}\label{subfigure:seg_noise_dilation}
    \end{subfigure}
    \\\vspace{1em}
    \begin{subfigure}{\subfigfrac\linewidth}
        {\includegraphics[width=\linewidth]{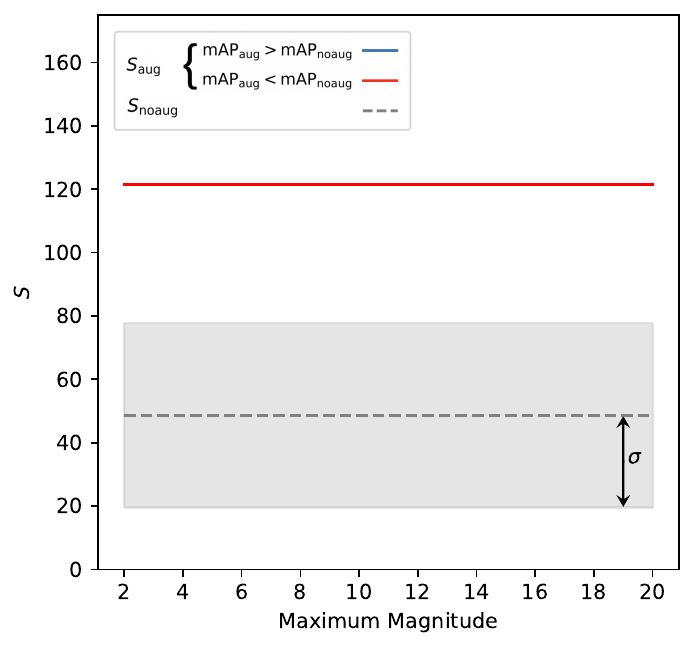}}\caption{}\label{subfigure:seg_noise_rectify}
    \end{subfigure}
    \begin{subfigure}{\subfigfrac\linewidth}
        {\includegraphics[width=\linewidth]{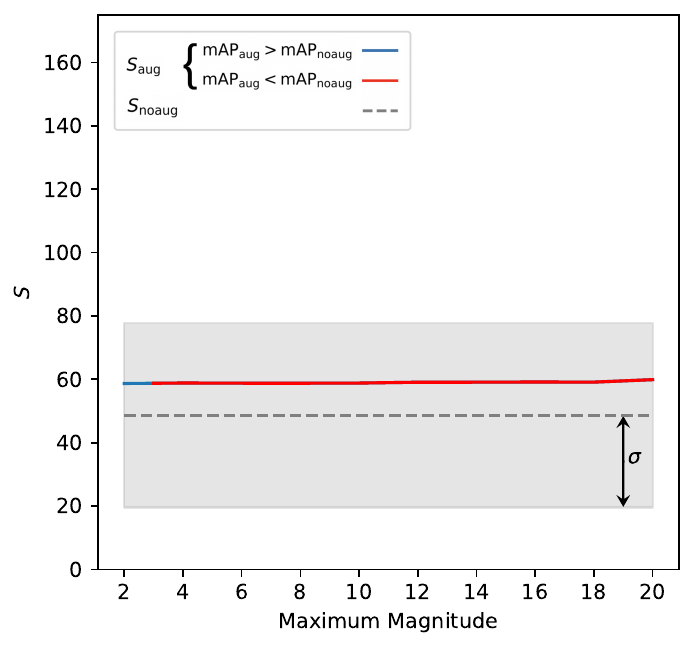}}\caption{}\label{subfigure:seg_noise_border_deform}
    \end{subfigure}
    \begin{subfigure}{\subfigfrac\linewidth}
        {\includegraphics[width=\linewidth]{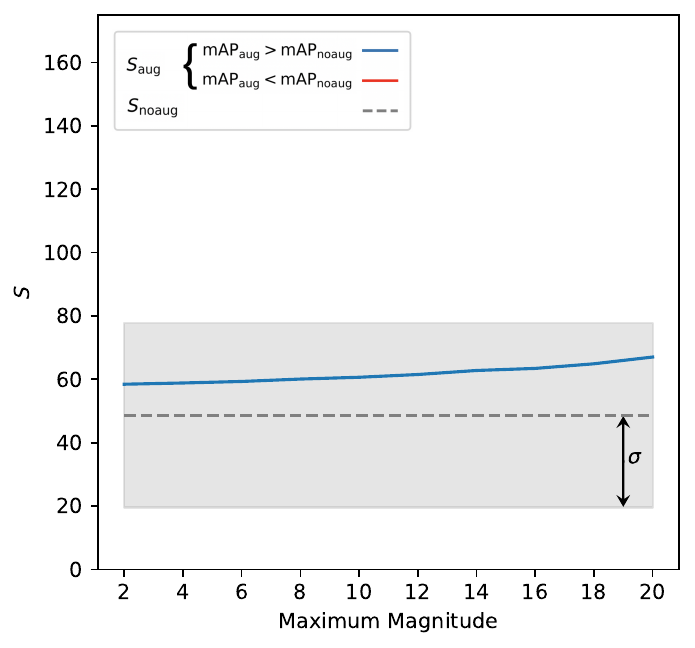}}\caption{}\label{subfigure:seg_noise_segment_swap}
    \end{subfigure}
    \begin{subfigure}{\subfigfrac\linewidth}
        {\includegraphics[width=\linewidth]{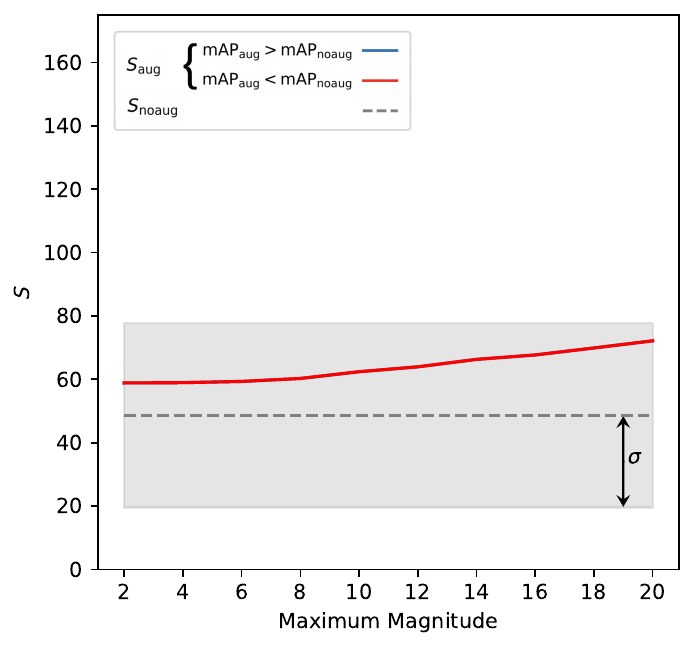}}\caption{}\label{subfigure:seg_noise_class_swap}
    \end{subfigure}
    \caption{\label{fig:results} Analysis of the degree of physical consistency of channel augmentation techniques. The score $S_{\text{noaug}}$ is plotted as the dashed gray line with its standard deviation $\sigma$ depicted in light gray. The solid line represents the score for $S_{\text{aug}}$. If the training performance of applying the respective augmentation technique with the corresponding maximum magnitude during training ($\text{mAP}_{\text{aug}}$) exceeds the training performance without applying any augmentation techniques ($\text{mAP}_{\text{noaug}}$), the solid line is depicted in blue. If the opposite holds true, the solid line is depicted in red. The considered augmentation techniques are: (a) Brightness; (b) Contrast; (c) Gaussian Blur; (d) Gaussian Noise; (e) Grayscale; (f) Posterize; (g) Sharpness; and (h) Solarize.}
\end{figure*}

\cref{fig:results} shows a comparison between the scores of unaugmented signatures $S_{\text{noaug}}$ and augmented signatures $S_{\text{aug}}$ for different channel augmentation techniques and maximum magnitudes. It can be observed that the stronger the magnitude of a channel augmentation technique, the larger the value for $S_{\text{aug}}$. Nonetheless, there are significant differences between the individual channel augmentation techniques. While the values $S_{\text{aug}}$ for contrast, Gaussian blur, Gaussian noise, posterize, sharpness and solarize stay within the standard deviation of $S_{\text{noaug}}$, brightness and grayscale seem to clearly affect the physical consistency by exceeding the expected deviation of signatures. While grayscale exceeds the standard deviation of the signature deviation of an original time series by $40$, the channel augmentation technique brightness surpasses the standard deviation at a maximum magnitude of 6 ($\sim$0.12 brightness factor), linearly increasing up to a difference of $80$ between $S_{\text{aug}}$ and the standard deviation of $S_{\text{noaug}}$ for a maximum magnitude of 20.

To relate the scores of expected deviations of augmented and unaugmented signatures with the improvements on the training of a neural network when leveraging the respective channel augmentation technique, we conduct further experiments. In detail, we train a neural network with each technique at each maximum magnitude as well as without applying augmentations. Therefore, we follow a standard setup for training a neural network on a multi-label classification task. We train a ResNet18 with a batch size of $512$ and an AdamW optimizer. Further, the optimizer is scheduled by a cosine annealing learning rate scheduling with a minimum and maximum learning rate of 1e-3 and 1e-4 for $30$ epochs. We conduct runs with $5$ different seeds and measure the final performance as an averaged score in mean average precision macro (mAP) on the test set. For each channel augmentation technique and maximum magnitude, we conduct experiments applying the respective augmentation technique with a probability of $50$\% during training (performance denoted as $\text{mAP}_{\text{aug}}$) and compare it to a baseline run that does not leverage any augmentation technique (performance denoted as $\text{mAP}_{\text{noaug}}$). The results are integrated into \cref{fig:results} by depicting the curve of $S_{\text{aug}}$ in blue if the channel augmentation technique at a given maximum magnitude yields an mAP score higher than the baseline, and in red if the run without augmentations yields the higher mAP score. Note that, on the one hand, if the score of $S_{\text{aug}}$ for an augmentation technique exceeds the standard deviation of $S_{\text{noaug}}$, the test performance of a training run leveraging this technique cannot improve a baseline as the physical consistency is affected. On the other hand, if the physical consistency of signatures is not affected by a channel augmentation technique (i.e., $S_{\text{aug}}$ stays within the standard deviation of $S_{\text{noaug}}$), whether this technique can improve over an augmentation-free baseline depends on the technique itself and not on the degree of physical consistency. Based on our analysis, we observe the following. Contrast and sharpness do not affect the physical consistency of signatures and almost constantly improve the training baseline (i.e., not applying any augmentation technique) when leveraged with any maximum magnitude. Gaussian blur and Gaussian noise can improve the training baseline. However, there is no clear trend whether stronger or weaker magnitudes consistently attain this effect. Posterize and solarize do not improve the baseline training, even though $S_{\text{aug}}$ indicates physical consistency. Finally, brightness above a maximum magnitude of 6 ($\sim$0.12 brightness factor) and grayscale affect the physical consistency as $S_{\text{aug}}$ exceeds the standard deviation of $S_{\text{noaug}}$. Consequently, training with these augmentation techniques can not improve compared to a baseline model without augmentations.

\section{Conclusion and Discussion}

In this paper, we have proposed an approach to measure the alignment of a pixel signature within a time series to provide a different perspective on the effects of channel augmentation techniques in the context of physical consistency of RS images. In detail, the proposed approach includes an estimation of scores for the expected values of deviation from the closest unaugmented pixel signature within a time series for (1) an unaugmented signature and (2) an augmented signature. Further, the comparison of the two scores is associated with the generalization capabilities of a deep neural network trained with and without the respective channel augmentation technique. Our results show that techniques such as contrast, Gaussian blur, Gaussian noise, posterize, sharpness, and solarize do not affect the physical consistency of the signatures as the scores for using these augmentation techniques stay within the standard deviation of the score for unaugmented signatures. In contrast, the usage of grayscale and brightness adjustments with a maximum magnitude of more than 6 ($\sim$0.12) exceeds the standard deviation and, thus, affects physical consistency. While physical inconsistency is a predictor for inferior generalization performance when leveraging the respective data augmentation technique during training, the opposite does not hold. For example, posterize and solarize do not affect the physical consistency more than the expected natural effects caused by the image acquisition conditions. Nonetheless, they can not improve a training procedure without augmentations. As a future work, we plan to extend our analysis to different modalities of RS data and analyze its effect in the context of self-supervised learning. 

\section{Acknowledgement}
This work is supported by the European Research Council (ERC) through the ERC-2017-STG BigEarth Project under Grant 759764 and by the European Space Agency through the DA4DTE (Demonstrator precursor Digital Assistant interface for Digital Twin Earth) project.

\printbibliography[heading=secbib, title=REFERENCES]

@inproceedings{sumbul_bigearthnet_2019,
	title = {Bigearthnet: {A} large-scale benchmark archive for remote sensing image understanding},
	booktitle = {{IEEE} {International} {Geoscience} and {Remote} {Sensing} {Symposium}},
	author = {Sumbul, Gencer and Charfuelan, Marcela and Demir, Begüm and Markl, Volker},
	year = {2019},
	pages = {5901--5904},
}

@inproceedings{he_momentum_2020,
	title = {Momentum contrast for unsupervised visual representation learning},
	booktitle = {Proceedings of the {IEEE}/{CVF} {Conference} on {Computer} {Vision} and {Pattern} {Recognition}},
	author = {He, Kaiming and Fan, Haoqi and Wu, Yuxin and Xie, Saining and Girshick, Ross},
	year = {2020},
	pages = {9729--9738},
}

@inproceedings{cubuk_autoaugment_2019,
	title = {{AutoAugment}: {Learning} augmentation strategies from data},
	booktitle = {Proceedings of the {IEEE}/{CVF} {Conference} on {Computer} {Vision} and {Pattern} {Recognition}},
	author = {Cubuk, Ekin D. and Zoph, Barret and Mane, Dandelion and Vasudevan, Vijay and Le, Quoc V.},
	year = {2019},
}

@article{sohn_fixmatch_2020,
	title = {Fixmatch: {Simplifying} semi-supervised learning with consistency and confidence},
	volume = {33},
	journal = {Advances in Neural Information Processing Systems},
	author = {Sohn, Kihyuk and Berthelot, David and Carlini, Nicholas and Zhang, Zizhao and Zhang, Han and Raffel, Colin A and Cubuk, Ekin Dogus and Kurakin, Alexey and Li, Chun-Liang},
	year = {2020},
	pages = {596--608},
}

@inproceedings{liu_convnet_2022,
	title = {A convnet for the 2020s},
	booktitle = {Proceedings of the {IEEE}/{CVF} {Conference} on {Computer} {Vision} and {Pattern} {Recognition}},
	author = {Liu, Zhuang and Mao, Hanzi and Wu, Chao-Yuan and Feichtenhofer, Christoph and Darrell, Trevor and Xie, Saining},
	year = {2022},
	pages = {11976--11986},
}

@inproceedings{lu_rsi-mix_2022,
	title = {{RSI}-{Mix}: {Data} augmentation method for remote sensing image classification},
	doi = {10.1109/ICSP54964.2022.9778421},
	booktitle = {7th {International} {Conference} on {Intelligent} {Computing} and {Signal} {Processing}},
	author = {Lu, Xingshun and Zhang, Chao and Ye, Qihong and Wang, Chao and Yang, Chuansheng and Wang, Quanqing},
	year = {2022},
	pages = {1982--1985},
}

@article{zhang_new_2021,
	title = {A new data augmentation method of remote sensing dataset based on class activation map},
	volume = {1961},
	doi = {10.1088/1742-6596/1961/1/012023},
	number = {1},
	journal = {Journal of Physics: Conference Series},
	author = {Zhang, Wei and Cao, Yungang},
	year = {2021},
	pages = {012023},
}

@article{burgert_effects_2022,
	title = {On the effects of different types of label noise in multi-label remote sensing image classification},
	volume = {60},
	doi = {10.1109/TGRS.2022.3226371},
	journal = {IEEE Transactions on Geoscience and Remote Sensing},
	author = {Burgert, Tom and Ravanbakhsh, Mahdyar and Demir, Begüm},
	year = {2022},
	pages = {1--13},
}

@article{stivaktakis_deep_2019,
	title = {Deep learning for multilabel land cover scene categorization using data augmentation},
	volume = {16},
	doi = {10.1109/LGRS.2019.2893306},
	number = {7},
	journal = {IEEE Geoscience and Remote Sensing Letters},
	author = {Stivaktakis, Radamanthys and Tsagkatakis, Grigorios and Tsakalides, Panagiotis},
	year = {2019},
	pages = {1031--1035},
}

@article{wang_ssl4eo-s12_2022,
	title = {{SSL4EO}-{S12}: {A} large-scale multi-modal, multi-temporal dataset for self-supervised learning in earth observation},
	journal = {arXiv preprint arXiv:2211.07044},
	author = {Wang, Yi and Braham, Nassim Ait Ali and Xiong, Zhitong and Liu, Chenying and Albrecht, Conrad M and Zhu, Xiao Xiang},
	year = {2022},
}

@article{patnala_generating_2023,
	title = {Generating views using atmospheric correction for contrastive self-supervised learning of multispectral images},
	volume = {20},
	doi = {10.1109/LGRS.2023.3274493},
	journal = {IEEE Geoscience and Remote Sensing Letters},
	author = {Patnala, Ankit and Stadtler, Scarlet and Schultz, Martin G. and Gall, Juergen},
	year = {2023},
	pages = {1--5},
}

@article{zhao_hyperspectral_2022,
	title = {Hyperspectral image classification with contrastive self-supervised learning under limited labeled samples},
	volume = {19},
	doi = {10.1109/LGRS.2022.3159549},
	journal = {IEEE Geoscience and Remote Sensing Letters},
	author = {Zhao, Lin and Luo, Wenqiang and Liao, Qiming and Chen, Siyuan and Wu, Jianhui},
	year = {2022},
	pages = {1--5},
}

@inproceedings{wang_self-supervised_2022,
	title = {Self-supervised vision transformers for joint {SAR}-optical representation learning},
	doi = {10.1109/IGARSS46834.2022.9883983},
	booktitle = {{IEEE} {International} {Geoscience} and {Remote} {Sensing} {Symposium}},
	author = {Wang, Yi and Albrecht, Conrad M and Zhu, Xiao Xiang},
	year = {2022},
	pages = {139--142},
}

@article{pan_coinnet_2019,
	title = {{CoinNet}: {Copy} {Initialization} network for multispectral imagery semantic segmentation},
	volume = {16},
	doi = {10.1109/LGRS.2018.2880756},
	number = {5},
	journal = {IEEE Geoscience and Remote Sensing Letters},
	author = {Pan, Bin and Shi, Zhenwei and Xu, Xia and Shi, Tianyang and Zhang, Ning and Zhu, Xinzhong},
	year = {2019},
	pages = {816--820},
}

@article{burgert_label_2024,
	title = {A {label} {propagation} {strategy} for {cutmix} in {multi}-{label} {remote} {sensing} {image} {classification}},
	journal = {arXiv preprint arXiv:2405.13451},
	author = {Burgert, Tom and Siebert, Tim and Clasen, Kai Norman and Demir, Begüm},
	year = {2024},
}
\end{document}